\documentclass[acmsmall]{acmart}

\usepackage{times}
\usepackage{soul}
\usepackage{url}
\usepackage[utf8]{inputenc}
\usepackage{graphicx}
\usepackage{amsmath}
\usepackage{amsthm}
\usepackage{booktabs}
\usepackage{algorithm}
\usepackage{algorithmic}
\usepackage[switch]{lineno}

\usepackage{amsmath}
\usepackage{bm}
\usepackage{enumitem}
\usepackage{xspace}
\usepackage{multirow}
\usepackage{adjustbox}
\usepackage{booktabs}
\DeclareUnicodeCharacter{2212}{-}

\usepackage{xcolor}
\usepackage{lipsum}
\usepackage[switch]{lineno}
\usepackage{bbding} 
\usepackage{color}

\newcommand{\ours}{MMICT\xspace}

\begin{document}

\title{MMICT: Boosting Multi-Modal Fine-Tuning with In-Context Examples}

\author{Tao Chen}
\authornote{Both authors contributed equally.}
\email{taochen@stu.xmu.edu.cn}
\affiliation{%
  \institution{Key Laboratory of Multimedia Trusted Perception and Efficient Computing, Ministry of Education of China, Xiamen University}
  \country{China}
}

\author{Enwei Zhang}
\authornotemark[1]
\email{miyozhang@tencent.com}
\author{Yuting Gao}
\email{yutinggao@tencent.com}
\author{Ke Li}
\email{tristanli.sh@gmail.com}
\author{Xing Sun}
\email{winfred.sun@gmail.com}
\affiliation{%
  \institution{Tencent Youtu Lab}
  \country{China}}

\author{Yan Zhang}
\email{bzhy986@gmail.com}
\author{Hui Li}
\authornote{Corresponding Author.}
\email{hui@xmu.edu.cn}
\author{Rongrong Ji}
\email{rrji@xmu.edu.cn}
\affiliation{%
  \institution{Key Laboratory of Multimedia Trusted Perception and Efficient Computing, Ministry of Education of China, Xiamen University}
  \country{China}
}

\renewcommand{\shortauthors}{Chen et al.}

\setcopyright{acmlicensed}
\copyrightyear{2024}
\acmYear{2024}

\acmJournal{ToMM}

\begin{abstract}
  Although In-Context Learning (ICL) brings remarkable performance gains to Large Language Models (LLMs), the improvements remain lower than fine-tuning on downstream tasks. This paper introduces Multi-Modal In-Context Tuning (MMICT), a novel multi-modal fine-tuning paradigm that boosts multi-modal fine-tuning by fully leveraging the promising ICL capability of multi-modal LLMs (MM-LLMs). We propose the Multi-Modal Hub (M-Hub), a unified module that captures various multi-modal features according to different inputs and objectives. Based on M-Hub, MMICT enables MM-LLMs to learn from in-context visual-guided textual features and subsequently generate outputs conditioned on the textual-guided visual features. Moreover, leveraging the flexibility of M-Hub, we design a variety of in-context demonstrations. Extensive experiments on a diverse range of downstream multi-modal tasks demonstrate that MMICT significantly outperforms traditional fine-tuning strategy and the vanilla ICT method that directly takes the concatenation of all information from different modalities as input. Our implementation is available at: https://github.com/KDEGroup/MMICT.
\end{abstract}

\ccsdesc[500]{Computing methodologies~Natural language generation}
\ccsdesc[500]{Computing methodologies~Computer vision tasks}

\keywords{Multi-Modal Alignment, Text Generation, In-Context Tuning}


\maketitle

\section{Introduction}

Recently, a great number of works on large-scale language models (LLMs)~\cite{abs-2303-18223} have sprung up, propelling the evolution of human-like artificial intelligence. By escalating the model size, for instance, from 1 billion (GPT 1) to 175 billion parameters (GPT 3) or more, LLMs can demonstrate extraordinary proficiency in comprehending human language. 
Many researchers attempt to further augment the text-based LLMs by incorporating additional modalities (e.g., image, and video), leading to the creation of multi-modal LLMs (MM-LLMs). 
Representative works include but not limited to KOSMOS-1~\cite{abs-2302-14045}, Flamingo~\cite{AlayracDLMBHLMM22} and GPT-4~\cite{abs-2303-08774}.

With the prosperous development, LLMs have shown a capacity for in-context learning (ICL)~\cite{abs-2301-00234}, which involves learning and prediction solely based on a few examples in the context and does not update any parameters of LLMs. 
For instance, in the field of MM-LLMs, Flamingo capitalizes on the interleaved multi-modal data to enhance its multi-modal in-context learning capabilities. 
Building upon OpenFlamingo~\cite{anas_awadalla_2023_7733589}, the open-source version of Flamingo, Otter~\cite{abs-2305-03726} is capable of 
executing new instructions with a few in-context learning examples using multi-modal in-context instruction tuning.
Additionally, some studies~\cite{ZhaoWFK021,LuBM0S22} verify that various demonstration factors (e.g., demonstration format and demonstration order) heavily affect the performance of ICL.

Although ICL can bring remarkable performance gains to MM-LLMs, the improvements still lag behind fine-tuning on training data for downstream tasks~\cite{AlayracDLMBHLMM22,ChenDPMISK22}. 
For example, on the VQAv2 task, Flamingo achieves 63.1\% accuracy with 4 demonstration examples while the accuracy after fine-tuning is 82.0\%.
The observation inspires us: Can we combine the two learning paradigms by leveraging ICL to further enhance the fine-tuning performance on downstream multi-modal tasks?

To this end, in this paper, we propose \underline{M}ulti-\underline{M}odal \underline{I}n-\underline{C}ontext \underline{T}uning (\ours), a novel multi-modal fine-tuning paradigm that harnesses ICL to improve multi-modal fine-tuning.
\ours enables MM-LLMs to learn from visual-guided textual features of demonstration examples in fine-tuning. 
Furthermore, based on the in-context information and the textual-guided visual features extracted from visual inputs and textual instructions, \ours predicts the textual label paired with the visual inputs.

\ours is built based on BLIP-2~\cite{abs-2301-12597}.
BLIP-2 adopts a traditional fine-tuning strategy with only query inputs, and exclusively utilizes the cross-modal pre-trained Qformer to extract visual features. 
To better capture the multi-modal features within a unified model architecture, we design the Multi-Modal Hub (M-Hub) used in \ours.
Different from Qformer, M-Hub can produce either uni-modal features or multi-modal features that fuse information from different modalities. 
Considering that different demonstration factors (e.g., feature extraction strategy, sampling number for demonstrations and sampling strategies for demonstrations) may heavily affect the performance, we design various variants of in-context demonstrations by leveraging the flexibility of M-Hub.

In summary, the contributions of this work are:
\begin{itemize}

    \item \textbf{Innovative Paradigm.} We introduce \ours, a novel fine-tuning paradigm that can further augment the performance of MM-LLMs on a variety of downstream multi-modal tasks by harnessing its promising ICL capability. Furthermore, our proposed \ours exhibits robustness against varying demonstration surfaces.

    \item \textbf{Thoughtful Design.} Based on the unique model architecture and representation learning strategy of Q-former used in the pre-training stage, we transcend Q-former's conventional use as a uni-modal feature extraction module by advancing it to M-Hub that is capable of capturing both uni-modal representations and visual-language representations within a unified architecture.

    \item \textbf{Insightful Discoveries.} Through the exploration of various demonstration formats, we unveil several intriguing and pivotal findings. These insights illuminate potential explanations and pave the way for future research in this domain.

\end{itemize}

The remaining parts of this paper are organized as follows:
Sec.~\ref{sec:related} introduces the related work of this study.
Sec.~\ref{sec:method} describes the details of \ours.
Sec.~\ref{sec:exp} provides the results and analysis of experiments.
Sec.~\ref{sec:con} concludes this work.

\section{Related Work}
\label{sec:related}

\subsection{Multi-Modal Large Language Models}

In recent years, the trend of using LLMs to integrate information from multiple modalities has gained significant attention, resulting in the so-called MM-LLMs.

Pioneering studies like VisualGPT~\cite{ChenGY0E22} and Frozen~\cite{TsimpoukelliMCE21} have demonstrated the benefits of employing a pre-trained language model as a vision-language model decoder. Flamingo~\cite{AlayracDLMBHLMM22} is proposed to align a pre-trained vision encoder and language model using the gated cross-attention mechanism.
It is trained on billions of image-text pairs, showcasing impressive in-context few-shot learning capabilities. BLIP-2~\cite{abs-2301-12597} introduces a Q-Former to efficiently align visual features with the language model. 
GPT-4~\cite{abs-2303-08774} shows more powerful visual understanding and reasoning abilities after pre-training on a vast collection of aligned image-text data. 
To empower LLMs with the ability of video understanding, a multi-branch cross-modal pre-training framework Video-LLaMA~\cite{abs-2306-02858} is proposed to achieve both vision-language alignment and audio-language alignment by connecting the LLM to off-the-shelf uni-modal pre-trained models.
MoE-LLaVA~\cite{abs-2401-15947} constructs a MoE-based sparse LVLM architecture with an outrageous number of parameters but a constant computational cost.
Gemini family~\cite{abs-2312-11805} exhibits remarkable capabilities across image, audio, video, and text understanding even on memory-constrained devices.

In summary, there are four mainstream methods that combine visual encoder and LLM into MM-LLM: 
1) The addition of extra modules in the LLM that enable deep interaction and fusion, exemplified by Otter~\cite{abs-2305-03726} and CogVLM~\cite{abs-2311-03079}, which incurs high computational costs. 
2) The incorporation of learnable query tokens to extract information in a query-based manner, facilitating the conversion of features to enhance the comprehension of LLM, as demonstrated in the BLIP-2~\cite{abs-2301-12597}, X-LLM~\cite{abs-2305-04160} and MiniGPT4~\cite{abs-2304-10592}, which compress visual tokens into a smaller number of representation vectors. 
3) The inclusion of supplementary learnable parameters in the LLM, as seen in the LLaMA-Adapter~\cite{abs-2304-15010}, yields a faster training speed albeit with marginally lower performance in comparison to alternative approaches.
4) The simple usage of a MLP-based interface to bridge the modality gap. For example, LLaVA series adopts one/two linear MLP~\cite{abs-2310-03744, LiuLWL23a} to project visual tokens and align the feature dimension with word embeddings. Despite their simplicity, the large amount of prefix visual tokens increases the context length in MM-LLM.
Considering the pros and cons of the aforementioned methods, we opt for BLIP-2 as the base of \ours.

\subsection{In-Context Learning}

\label{rw_icl}

In-context learning involves learning based on only a few examples in the form of demonstration. 
Essentially, it estimates the likelihood of the potential answer conditioned on the demonstration using a well-trained language model. 
For multi-modal tasks, Flamingo~\cite{AlayracDLMBHLMM22} capitalizes on the interleaved multi-modal data to enhance its few-shot ICL capabilities. 
Moreover, the paradigm of generating query text conditioned on in-context examples ensures its ICL capacity during the inference phase. 
Building upon OpenFlamingo, Otter~\cite{abs-2305-03726} introduces the in-context instruction tuning paradigm for multi-modal models.
However, they do not capture cross-modality information and only use uni-modal features from different modalities as demonstrations. 
The multi-modal features are modeled in the cross-attention module of the LLM.
Differently, \ours models cross-modality information to guide the construction of the multi-modal demonstrations. More detailed comparisons are provided in Sec.~\ref{cmp_otter}.

As demonstrations play a vital role in ICL, many works study demonstration designing strategies. 
For instance, in natural language processing, several works aim to select good examples for ICL through unsupervised methods based on pre-defined metrics~\cite{abs-2302-13539} or supervised methods~\cite{abs-2301-11916}.
For tasks requiring complex reasoning (e.g., math word problems and commonsense reasoning), some works design better demonstrations for ICL by describing tasks with the instruction~\cite{WangKMLSKH23} and adding intermediate reasoning steps~\cite{abs-2305-08848}. 
Differently, \ours proposes a better feature extraction strategy for demonstrations to avoid information redundancy and exhibits robustness against different demonstration surfaces.

\subsection{Multi-modal Alignment}

Multi-modal tasks necessitate a model that adeptly translates visual scenes into language descriptions.
Currently popular approaches~\cite{ShiXY0HZ23, LiFPYLM22, DongN0LX23} leverage an encoder-decoder architecture and achieve promising results. 
An optimal multi-modal model embodies the trifecta of visual comprehension, visual-text alignment, and textual generation.
Initially, a diverse array of feature extractors is employed to procure visual representations, fundamental for basic visual understanding.
Subsequently, these visual representations are transformed into a hidden vector $h$, which shares a semantic space with the language representations' hidden vector, thus bridging the gaps between vision and language.
In essence, this alignment occurs in a shared semantic space, uniting the visual modality (e.g., video or image) with the textual modality. Ultimately, the language representations undergo decoding to produce the definitive descriptions.
The pivotal step in this multi-modal model is the visual-text alignment, mapping visual representations to the language domain.
This alignment is crucial because the language network (such as LSTM or Transformer) within the decoder can unleash its full text generation prowess when the input representation resonates with the language domain.
However, existing techniques tend to concentrate either on video comprehension~\cite{PengSSZ24, TangWL19} or image understanding~\cite{WangYM18, JiangWH21}. Thanks to the flexibility of our proposed M-Hub, \ours is capable of executing text generation tasks, whether based on images or videos, within a unified framework.
\section{Our Method}
\label{sec:method}

\begin{figure*}[t]
    \centering
    \includegraphics[width=0.98\textwidth]{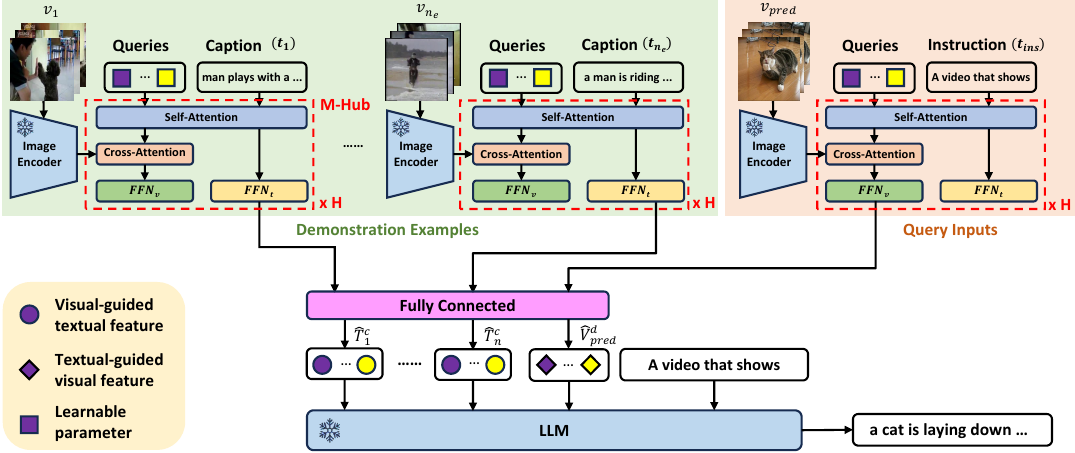}
    \caption{Overview of \ours. M-Hub can output both visual-guided textual features (upper left green part) and instruction-guided visual features (upper right orange part). MMICT learns from visually-guided textual features derived from demonstration examples and generates outputs based on instruction-guided visual features obtained from input queries.}
    \label{fig:overview}
\end{figure*}

In this section, we illustrate the details of \ours. Fig.~\ref{fig:overview} provides an overview of \ours.
Firstly, we demonstrate how \ours learns from in-context visual-guided textual features and generates outputs according to textual-guided visual features in Sec.~\ref{sec:ict}. 
Then, in Sec.~\ref{sec:mencoder}, we illustrate the Multi-Modal Hub (M-Hub) that encodes the multi-modal fused features in \ours within a unified architecture.
In the following, we use lower-case fonts to indicate raw data, bold lower-case fonts to represent vectors, and bold upper-case fonts to denote  matrices.

\subsection{Multi-Modal In-Context Tuning}
\label{sec:ict}

For multi-modal tasks, \ours introduces a novel fine-tuning paradigm that fully leverages the remarkable ICL capacity of MM-LLMs. 
If only the visual or textual data is selected and fed into MM-LLMs, it could lead to a performance decline due to the absence of information from the other modality. 
To encapsulate the visual-textual information present in multi-modal tasks, a straightforward method could be directly concatenating the visual and textual data from the demonstration examples together, and then feeding them into MM-LLMs.
This simplistic approach, however, is suboptimal as it incorporates a significant amount of redundant information from the visual and textual modalities. 
Instead, we argue that fusing multi-modal information as demonstrations is a more effective strategy, as it not only integrates information from different modalities but also circumvents information redundancy.

In this study, we feed paired features from different modalities into the Multi-modal Hub (M-Hub, see Sec.~\ref{sec:mencoder}) to obtain multi-modal fused features. 
Moreover, considering the modality gap between vision and text, we retain the visual-guided textual features as demonstrations. 
To elucidate the detailed formulation of \ours, let us take video captioning as an example. 
As depicted in Fig.~\ref{fig:overview}, given a video-instruction pair $\{v_{pred}, t_{ins}\}$ (upper right orange part in Fig.~\ref{fig:overview}) accompanied with other pairs $\{v_{1}, t_{1},...,v_{n_{e}}, t_{n_{e}}\}$ (upper left green part in Fig.~\ref{fig:overview}) that are randomly selected as demonstration examples, where $\{v_{*}, t_{*}\}$ denotes a video clip and the corresponding text, and $n_{e}$ is the number of demonstration examples. 
The model needs to predict the label $y$ (e.g., the caption text ``a cat is laying down washing his face'') paired with $v_{pred}$ according to the in-context information.
The frozen image encoder (e.g. EVA~\cite{abs-2211-07636}) takes as input the video clips $\{v_{1},...,v_{n_{e}},v_{pred}\}$ and outputs the corresponding encoded visual features $\{\mathbf{Z}^{v}_{1},...,\mathbf{Z}^{v}_{n_{e}},\mathbf{Z}^{v}_{pred}\}$.
We first feed these features and their paired text into M-Hub, and then pass them to a share-weight fully-connected network to extract visual-guided textual demonstration features and textual-guided visual features: 
\begin{equation}
    \label{eq:vg_t}
    \begin{aligned}
    &\mathbf{\hat{T}}^{c}_{k} = \text{FC}\big(\mathcal{G}(\mathbf{Z}_{k}^{v}, t_{k})\big) \\
    \end{aligned}
\end{equation}
\begin{equation}
    \label{eq:tg_v}
    \begin{aligned}
    &\mathbf{\hat{V}}^{d}_{pred} = \text{FC}\big(\mathcal{G}(\mathbf{Z}_{pred}^{v}, t_{ins})\big)
    \end{aligned}
\end{equation}
where $k \in \{1,...,n_{e}\}$. $\text{FC}$ and $t_{ins}$ denote
single-layer fully-connected network and textual instruction, respectively.
$\mathcal{G}$ indicates M-Hub.
Finally, we concatenate multi-modal fused features with $t_{ins}$ and feed them into the frozen LLM:
\begin{equation}
    \label{eq:icl}
    \begin{aligned}
    &\mathbf{C} = \big[\mathbf{\hat{T}}^{c}_{1},\left \langle \text{EOC}\right \rangle,...,\mathbf{\hat{T}}^{c}_{n_{e}},\left \langle \text{EOC}\right \rangle,\mathbf{\hat{V}}^{d}_{pred}\big] \\
    &\hat{y} = \text{LLM}\big([\mathbf{C},\mathcal{T}(t_{ins})]\big) \\ 
    \end{aligned}
\end{equation}
where $[\cdot,\cdot]$ denotes the concatenation operation. 
$\left \langle \text{EOC}\right \rangle$ token (``end of chunk'') is appended to the end of each demonstration example for separating them explicitly, and $\mathcal{T}$ is the textual tokenizer from the LLM.
$\hat{y}$ denotes the outputs of the model conditioned on the in-context information, and it gradually gets closer to the ground truth $y$ during training.

\subsection{Multi-Modal Hub}
\label{sec:mencoder}

Owing to the huge model size of LLMs and vision foundation models, training their parameters for multi-modal tasks proves to be challenging. 
As a result, many researchers endeavor to incorporate a comparatively lightweight and trainable Visual Prompt Generator (VPG) between them while maintaining their fully frozen state~\cite{abs-2305-01278}. 
Among these efforts, the BLIP-style multi-modal pre-training approach effectively connects LLMs to vision foundation models via the Q-former. 
However, previous works~\cite{abs-2304-10592,abs-2301-12597} solely utilize this module to extract uni-modal visual features for LLMs after pre-training, thereby overlooking the benefits of multi-modal pre-training.

To address this issue, we propose the M-Hub for capturing multi-modal features. As depicted in the upper left part of the Fig.~\ref{fig:overview}, M-Hub comprises of $H$ blocks.
Specifically, each block in the M-Hub consists of a shared self-attention module, a cross-attention module that interacts with the frozen image encoder for extracting text-aligned visual features digestible for LLMs, and two modality-specific feed-forward layers. 
The queries are a set of learnable parameters $\mathbf{Q}=\{\mathbf{q}_{i}\}_{i=1}^{n_{q}}$, where $n_{q}$ is the number of parameters. 
We initialize the weights of the M-Hub using the pre-trained Q-former to leverage the advantages of multi-modal pre-training.

Given a video-text pair $\{v, t\}$, we uniformly sample $n_{f}$ frames from the video clip $v$, and subsequently extract frame features $\mathbf{Z}^{v}=\{\mathbf{E}^{v}_{i}\}_{i=1}^{n_{f}}$ by separately passing each frame through the frozen image encoder. 
Then, a simple approach can be acquiring frame-level visual features via individually feeding frame features into VPG and then concatenating them.
Considering that there exists both inter-modality information redundancy and intra-modality information redundancy (e.g., within a video clip), the aforementioned approach is suboptimal. 
To handle this problem, we propose a simple yet efficient method that fully capitalizes on its robust capacity to filter redundant information. 
Specifically, we flatten frame features and then feed them into M-Hub to obtain the video-level visual features.

Moreover, the unified architecture of the M-Hub can enhance the representation learning of visual-textual inputs. 
For instance, M-Hub can output both visual-guided textual features $\mathbf{\hat{T}}^{c}$ and textual-guided visual features $\mathbf{\hat{V}}^{d}$ as follows:
\begin{equation}
    \label{eq:mhub}
    \begin{aligned}
    &\mathbf{P}_{h}, \mathbf{R}_{h} = \text{Self-Attention}\big([\mathbf{\hat{V}}^{d}_{h-1}, \mathbf{\hat{T}}^{c}_{h-1}]\big) \\
    &\mathbf{O}_{h} = \text{Cross-Attention}\big(\mathbf{P}_{h}, \mathbf{Z}^{v}\big) \\
    &\mathbf{\hat{V}}^{d}_{h} = \text{FFN}_{v}\big(\mathbf{O}_{h}\big),\,\,\,\mathbf{\hat{T}}^{c}_{h} = \text{FFN}_{t}\big(\mathbf{R}_{h}\big)
    \end{aligned}
\end{equation}
where $\mathbf{\hat{V}}^{d}_{0}=\mathbf{Q}$, $\mathbf{\hat{T}}^{c}_{0}=t$, and $h$ is the $h$-th block of M-Hub. 
In the $h$-th block of M-Hub, we handle features $\mathbf{\hat{V}}^{d}_{h-1}$ and $ \mathbf{\hat{T}}^{c}_{h-1}$ from different modalities, which are outputs from the preceding block. To integrate multi-modal information, these features are concatenated within the self-attention layer. 
We partition the previous $n_{q}$ tokens in the concatenated features to derive multi-modal fused visual features $\mathbf{P}_{h}$, and retain the remaining tokens to obtain multi-modal fused textual features $\mathbf{R}_{h}$.
The valuable visual information $\mathbf{P}_{h}$ is extracted together with $\mathbf{Z}^{v}$ via the cross-attention layer, resulting in $\mathbf{O}_{h}$. Finally, $\mathbf{O}_{h}$ and $\mathbf{R}_{h}$ are passed through modality-specific feed-forward layers to yield $\mathbf{\hat{V}}^{d}_{h}$ and $\mathbf{\hat{T}}^{c}_{h}$, respectively.
Note that, depending on the varying inputs and objectives, M-Hub can output other types of features. 
We will elaborate on its flexibility in Sec.~\ref{sec:icldv}.

\subsection{Training Objective}

\ours is trained on the next-token prediction task, i.e., learn to generate the next token depending on the previous context. 
The training objective is to minimize the negative log-likelihood of tokens in the label $y$:
\begin{equation}
    \mathcal{L} = -\sum_{y\in Y}\sum_{t=1}^{\left|y\right|} log\big(p(y_{t}|\mathbf{C},t_{ins},y_{1},...,y_{t-1})\big)
\end{equation}
where $y_t$ is the $t$-th token in the ground truth label $y$, $\left|y\right|$ is the number of tokens in $y$, and $Y$ is the ground truth label set.

\subsection{Inference}
During inference, \ours generates predictions as follows:
\begin{equation}
    \label{eq:inf}
    \hat{y} = \text{LLM}\big([\mathbf{\hat{V}}^{d}_{pred},\mathcal{T}(t_{ins})]\big)
\end{equation}
For evaluation, unlike the training stage, we first generate the predicted label $\hat{y}$.
Then, $\hat{y}$ is compared with the ground-truth label $y$ for calculating different evaluation metrics.

\section{Experiments}
\label{sec:exp}

\subsection{Implementation Details}

We follow BLIP-2\footnote{\url{https://github.com/salesforce/lavis}} to implement \ours. Concretely, we experiment with two LLMs: FlanT5~\cite{abs-2210-11416} with encoder-decoder architecture and OPT~\cite{abs-2205-01068} with decoder-only architecture.
In our approach, we utilize $\text{FlanT5}_{\text{XL}}$ for FlanT5 and $\text{OPT}_{\text{2.7B}}$ for OPT, respectively. 
For the frozen image encoder, we use EVA~\cite{abs-2211-07636}, a state-of-the-art pre-trained vision transformer model.
We initiate the parameters of M-Hub with the pre-trained Q-former.
We freeze the image encoder and the LLM, and only train the M-Hub and the fully-connected network (Eq.~\ref{eq:vg_t} and Eq.~\ref{eq:tg_v}) for better evaluating the effectiveness of \ours. 

\begin{table}[]
\centering
\caption{Instruction templates used for image captioning and video captioning.}
\label{tab:template}
\resizebox{0.6\columnwidth}{!}{
\begin{tabular}{@{}c|l@{}}
\toprule
Task                                                                        & Instruction Templates                                         \\ \midrule
\multirow{6}{*}{\begin{tabular}[c]{@{}c@{}}Image\\ Captioning\end{tabular}} & A short image caption:                                        \\
                                                                            & A image that shows                                            \\
                                                                            & Write a short description for the image.                      \\
                                                                            & Briefly describe the content of the image.                    \\
                                                                            & Use a few words to illustrate what is happening in the image. \\
                                                                            & Can you briefly explain what you see in the image?            \\ \midrule
\multirow{6}{*}{\begin{tabular}[c]{@{}c@{}}Video\\ Captioning\end{tabular}} & A short video caption:                                        \\
                                                                            & A video that shows                                            \\
                                                                            & Write a short description for the video.                      \\
                                                                            & Briefly describe the content of the video.                    \\
                                                                            & Use a few words to illustrate what is happening in the video. \\
                                                                            & Can you briefly explain what you see in the video?            \\ \bottomrule
\end{tabular}
}
\end{table}

\subsection{Evaluation Tasks and Datasets}

We evaluate \ours on several prevalent downstream multi-modal tasks, including image captioning, video captioning, visual question answering (VQA) and video question answering (VideoQA) across six different datasets.

For captioning tasks, we evaluate \ours on 3 public datasets including COCO Caption~\cite{LinMBHPRDZ14}, MSVD~\cite{XuZX0Z0Z17}, MSR-VTT~\cite{XuMYR16}. We use BLEU@4 (B@4)~\cite{PapineniRWZ02}, CIDEr (C)~\cite{VedantamZP15} as metrics. 
$t_{ins}$ is randomly sampled from the pre-defined instruction templates
, which are shown in Tab.~\ref{tab:template}

For open-ended question answering tasks, we evaluate \ours on 3 public datasets including VQAv2~\cite{GoyalKSBP17}, MSVD~\cite{XuZX0Z0Z17} and MSR-VTT~\cite{XuZX0Z0Z17}. 
We formulate them as generative problems.  
During inference, we use beam search with a beam width of 5 to generate answers from the whole vocabulary with no restrictions. 
Accuracy (Acc) is used as the evaluation metric. We evaluate VQAv2 on its validation set (the test label is not publicly available) and evaluate MSVD and MSR-VTT on their respective test sets.
$t_{ins}$ for these tasks is designed as ``Question: \{\} Answer:''.

\subsection{Environment and Hyper-parameters}
We follow most of the settings for fine-tuning hyper-parameters in BLIP-2, except that we freeze the image encoder and set the size of images/videos to be $224\times224$. 
We train our model for 5 epochs on 4 NVIDIA V100 GPUs.
Each video clip comprises of $n_{f}$ frames, we pass them through M-Hub to obtain video-level visual features. Therefore, we define the batch size in terms of the number of frames, and set the batch size on video tasks to $48$.
We treat an image as a single-frame video, where $n_{f}=1$, and use a batch size of 15 for image tasks.
$n_{f},n_{e},$ and $n_{q}$ are set to 16, 2, and 32, respectively. 
The demonstration examples for each data sample are randomly sampled from the same dataset.
The AdamW optimizer with $\beta_{1} = 0.9, \beta_{2} = 0.999$, and a weight decay of $0.05$ is used. 
Additionally, we apply a linear warmup of the learning rate during the initial 1,000 steps, increasing from $10^{−8}$ to $10^{−5}$, followed by a cosine decay with a minimum learning rate of $0$.

\subsection{Baselines}

For simplicity, we denote the formulation of \ours as $\{\mathbf{\hat{T}}^{c}_{1},...,\mathbf{\hat{T}}^{c}_{n_{e}},\mathbf{\hat{V}}^{d}_{pred}\}$, which represents the inputs to the LLM. 
We consider two baselines in our experiments:
\begin{itemize} 

    \item \textbf{VanillaFT}: It is the traditional method for fine-tuning on downstream tasks. Its formulation can be symbolized as $\{\mathbf{V}^{a}_{pred}\}$.

    \item \textbf{VanillaICT-$\text{B}_{\text{VT}}$}: VanillaICT denotes that M-Hub serves only as a uni-modal encoder. And we use `Base' (B) to indicate that the text is directly fed into the LLM. VanillaICT-$\text{B}_{\text{VT}}$ directly prompts an MM-LLM with the concatenation of all uni-modal information from demonstration examples to capture in-context information. We denote its formulation as $\{\mathbf{V}^{a}_{1},t_{1},...,\mathbf{V}^{a}_{n_{e}},t_{n_{e}},\mathbf{V}^{a}_{pred}\}$.

\end{itemize}

We mainly compare \ours with the above two baselines to demonstrate the superiority of using in-context learning to boost the fine-tuning performance of MM-LLMs. 
Additionally, we show the results of several state-of-the-art methods (SOTA) on each downstream task:
\begin{itemize} 

    \item \textbf{VLAB}~\cite{abs-2305-13167}: VLAB is a video language pre-training method that transfers CLIP’s learned representations to video-text tasks.

    \item \textbf{VAST}~\cite{abs-2305-18500}: VAST is an omni-modality video-text foundational model that can perceive and process vision, audio, and subtitle modalities from videos.

    \item \textbf{mPLUG-2}~\cite{abs-2302-00402}: mPLUG-2 is a multi-module composition network. It contains shared modules for modality collaboration and uses different modality modules to deal with modality entanglement.

\end{itemize}

\subsection{Overall Performance}

Tab.~\ref{tb:overall} reports the performance of \ours compared with two baselines, i.e., VanillaFT and VanillaICT-$\text{B}_{\text{VT}}$. 
The performance of SOTA methods is denoted in gray.
From the results shown in Tab.~\ref{tb:overall}, we can observe that:
\begin{enumerate} 

\item \ours outperforms baselines for four downstream multi-modal tasks on six datasets, and even achieves new SOTA results on MSVD for video captioning and VideoQA tasks. The results show that in-context tuning can enhance the performance of MM-LLMs on downstream multi-modal tasks.

\item A notable performance gap is observed between VanillaICT-$\text{B}_{\text{VT}}$ with OPT and other methods across most datasets. One possible explanation could be that the superfluous information from demonstration examples may considerably impact the performance of MM-LLMs with decoder-only architectures. In these architectures, the outputs are invariably influenced by the inputs via the mask self-attention module. Conversely, \ours utilize the in-context visual-guided textual features to incorporate multi-modal fused information and circumvent redundancy. This strategy consequently leads to a substantial enhancement in performance.

\end{enumerate}

\subsection{The Impacts of Demonstration Formats}
\label{sec:icldv}

\begin{table*}[]
\centering
\caption{Performance of all methods. 
We mainly compare our \ours with two baselines. The performance of SOTA methods is denoted in gray. The best results across different types of MM-LLMs are shown in bold.}
\resizebox{0.9\textwidth}{!}{
\begin{tabular}{ccccccccccc}
\toprule[2pt]
                         &                                   & \multicolumn{6}{c}{\textbf{Caption}}                                                                                                                                                         & \textbf{VQA}                      & \multicolumn{2}{c}{\textbf{VideoQA}}                               \\ \cmidrule(r){3-8} \cmidrule(r){9-9} \cmidrule(r){10-11}
                         &                                   & \multicolumn{2}{c}{\textbf{COCO}}                                   & \multicolumn{2}{c}{\textbf{MSR-VTT}}                               & \multicolumn{2}{c}{\textbf{MSVD}}                                   & \textbf{VQAv2}                    & \textbf{MSR-VTT}                     & \textbf{MSVD}                        \\ \cmidrule(r){3-4} \cmidrule(r){5-6} \cmidrule(r){7-8} \cmidrule(r){9-9} \cmidrule(r){10-10} \cmidrule(r){11-11}
\multirow{-3}{*}{\textbf{Method}}    & \multirow{-3}{*}{\textbf{LLM}}          & B@4                         & C                            & B@4                         & C                           & B@4                         & C                            & Acc                      & Acc                        & Acc                        \\ \midrule
\multicolumn{2}{c}{{\color[HTML]{C0C0C0} VLAB~\cite{abs-2305-13167}}}              & {\color[HTML]{C0C0C0} -}    & {\color[HTML]{C0C0C0} 152.5} & {\color[HTML]{C0C0C0} 54.6} & {\color[HTML]{C0C0C0} 74.9} & {\color[HTML]{C0C0C0} 79.3} & {\color[HTML]{C0C0C0} 179.8} & {\color[HTML]{C0C0C0} -} & {\color[HTML]{C0C0C0} 49.6} & {\color[HTML]{C0C0C0} 61.0} \\
\multicolumn{2}{c}{{\color[HTML]{C0C0C0} VAST~\cite{abs-2305-18500}}}              & {\color[HTML]{C0C0C0} -}    & {\color[HTML]{C0C0C0} 149.0} & {\color[HTML]{C0C0C0} -}    & {\color[HTML]{C0C0C0} 78.0} & {\color[HTML]{C0C0C0} -}    & {\color[HTML]{C0C0C0} -}     & {\color[HTML]{C0C0C0} -} & {\color[HTML]{C0C0C0} 50.1} & {\color[HTML]{C0C0C0} 60.2} \\
\multicolumn{2}{c}{{\color[HTML]{C0C0C0} mPLUG-2~\cite{abs-2302-00402}}}           & {\color[HTML]{C0C0C0} 41.6} & {\color[HTML]{C0C0C0} 137.7} & {\color[HTML]{C0C0C0} 57.8} & {\color[HTML]{C0C0C0} 80.3} & {\color[HTML]{C0C0C0} 75.0} & {\color[HTML]{C0C0C0} 165.8} & {\color[HTML]{C0C0C0} -} & {\color[HTML]{C0C0C0} 48.0} & {\color[HTML]{C0C0C0} 58.1} \\ \midrule
VanillaFT           & \multirow{3}{*}{FlanT5}              & 42.4       & 144.5       & 51.3                        & 74.7                        & 73.0                        & 174.3                        & 69.6                     & 43.4                        & 63.9                        \\
VanillaICT-$\text{B}_{\text{VT}}$   &   & 42.4                        & 142.5                        & 50.6                        & 74.4                        & 75.4                        & 178.5                        & 69.8                     & 42.8                        & 64.3                        \\
\textbf{\ours}     &                  & \textbf{43.6}                        & \textbf{145.7}                        & \textbf{51.6}                        & \textbf{74.8}                        & \textbf{76.5}                        & \textbf{179.1}                        & \textbf{72.3}            & \textbf{45.6}                        & \textbf{66.7}               \\ \midrule
VanillaFT           & \multirow{3}{*}{OPT}              & 43.7       & \textbf{145.8}       & 47.6                        & 69.9                        & 80.3                        & 177.1                        & 56.9                     & 42.6                        & 56.9                        \\
VanillaICT-$\text{B}_{\text{VT}}$  &.    & 36.3                        & 116.9                        & 37.7                        & 57.3                        & 43.2                        & 82.9                         & 62.7                     & 40.5                        & 64.7                        \\
\textbf{\ours}      &                  & \textbf{43.9}               & 145.5                        & \textbf{52.0}                        & \textbf{71.4}                        & \textbf{80.4}               & \textbf{180.4}               & \textbf{73.0}            & \textbf{46.0}                        & \textbf{66.3}                        \\ \bottomrule[2pt]
\end{tabular}
}
\label{tb:overall}
\end{table*}

\begin{figure*}[t]
    \centering
    \includegraphics[width=1\textwidth]{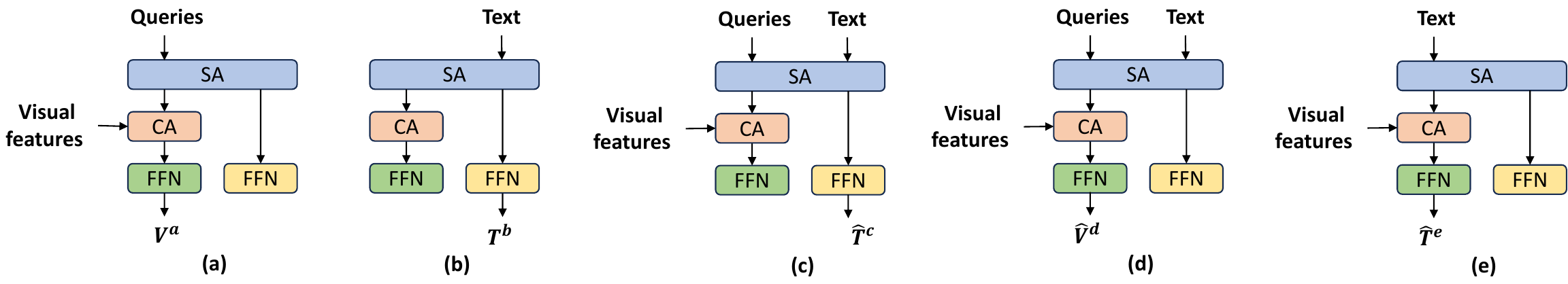}
    \caption{Different usages of M-Hub. As demonstrated in (a) and (b), it can function as a uni-modal encoder. Moreover, it can also operate as a multi-modal fusion encoder, as shown in (c), (d), and (e).}
    \label{fig:mmfe}
\end{figure*}

\begin{table}[]
\centering
\caption{The results of ablation studies about ICT demonstration variants. Best results are shown in bold. For the evaluation, we train on one third data of the complate datasets that is randomly sampled.}
\resizebox{0.65\textwidth}{!}{
\begin{tabular}{ccccccc}
\toprule
\multirow{3}{*}{Method}           & \multicolumn{4}{c}{Caption}                                     & VQA           & VideoQA       \\ \cmidrule(r){2-5} \cmidrule(r){6-6} \cmidrule(r){7-7}
                                   & \multicolumn{2}{c}{COCO}       & \multicolumn{2}{c}{MSVD}       & VQAv2         & MSVD          \\ \cmidrule(r){2-3} \cmidrule(r){4-5} \cmidrule(r){6-6} \cmidrule(r){7-7}
                                   & B@4           & C              & B@4           & C              & Acc           & Acc          \\ \midrule
VanillaICT-$\text{B}_{\text{VT}}$  & 40.6          & 138.6          & 73.7          & 175.9          & 69.0          & 62.5          \\
VanillaICT-$\text{B}_{\text{T}}$   & 41.8          & 140.2 & 74.7          & 176.7          & 68.4          & 61.5          \\
VanillaICT-$\text{E}_{\text{T}}$   & 41.9 & 140.0          & 75.0          & 177.5          & 68.8          & 63.5          \\ \midrule
InstructICT-$\text{E}_{\text{VT}}$ & 41.4          & 138.9          & 73.8          & 175.2          & 69.0          & 62.1          \\
InstructICT-$\text{E}_{\text{V}}$  & 38.7          & 133.4          & 69.7          & 166.3          & 69.4          & 62.9          \\
InstructICT-$\text{E}_{\text{T}}$  & 41.5          & 139.8          & 74.2          & 176.8          & 69.8          & 62.3          \\ \midrule
\ours                        & \textbf{42.0} & \textbf{140.4} & \textbf{76.2} & \textbf{177.7} & \textbf{69.9} & \textbf{64.3} \\ \bottomrule
\end{tabular}
}
\label{tb:ab_ICT}
\end{table}

As illustrated in Fig.~\ref{fig:mmfe}, depending on the different inputs and objectives, M-Hub can function as: 
\begin{enumerate}[label=(\alph*)] 
\item An uni-modal visual encoder. It can take image/video features as input, and output uni-modal visual features $\mathbf{V}^{a}$.

\item An uni-modal textual encoder that can output uni-modal textual features $\mathbf{T}^{b}$.

\item A multi-modal fusion encoder that can output visual-guided textual features $\mathbf{\hat{T}}^{c}$, where the learned queries and textual features interact with each other in the self-attention layers of the M-Hub.

\item A multi-modal fusion encoder that can capture textual-guided visual features $\mathbf{\hat{V}}^{d}$.

\item  Moreover, we replace the learned queries with input text to obtain the visual-attended textual features $\mathbf{\hat{T}}^{e}$ for exploring the performance of multi-modal representations obtained after direct interaction between visual features and textual features.

\end{enumerate}

Based on the flexibility of M-Hub mentioned above, we provide ablation studies to analyze the key factors that contribute to \ours’s performance, with insights and qualitative results. 
In InstructICT, M-Hub functions as a multi-modal fusion encoder. Additionally, we use `Encoding' (E) to signify that features from specific modalities are encoded by M-Hub.
We design various in-context demonstration variants as follows:

\begin{itemize} 

    \item \textbf{VanillaICT-$\text{B}_{\text{T}}$}: In VanillaICT-$\text{B}_{\text{T}}$, we remove the visual information from the demonstration examples to explore the information redundancy existed in VanillaICT-$\text{B}_{\text{VT}}$. We formulate VanillaICT-$\text{B}_{\text{T}}$ as $\{t_{1},...,t_{n_{e}},\mathbf{V}^{a}_{pred}\}$.

    \item \textbf{VanillaICT-$\text{E}_{\text{T}}$}: As shown in Fig.~\ref{fig:mmfe}~$(b)$, the M-Hub can work as the uni-modal text encoder. To explore its effectiveness, we design VanillaICT-$\text{E}_{\text{T}}$, which can be formulated as $\{\mathbf{T}^{b}_{1},...,\mathbf{T}^{b}_{n_{e}},\mathbf{V}^{a}_{pred}\}$.

    \item \textbf{InstructICT-$\text{E}_{\text{VT}}$}: We extend VanillaICT-$\text{B}_{\text{VT}}$ to InstructICT-$\text{E}_{\text{VT}}$, where the M-Hub works as the multi-modal fusion encoder. Its formulation can be denoted as $\{\mathbf{\hat{V}}^{d}_{1},\mathbf{\hat{T}}^{c}_{1},...,\mathbf{\hat{V}}^{d}_{n_{e}},\mathbf{\hat{T}}^{c}_{n_{e}},\mathbf{\hat{V}}^{d}_{pred}\}$.

    \item \textbf{InstructICT-$\text{E}_{\text{V}}$}: Compared with \ours, InstructICT-$\text{E}_{\text{V}}$ only retain the textual-guided visual features. Its formulation can be symbolized as $\{\mathbf{\hat{V}}^{d}_{1},...,\mathbf{\hat{V}}^{d}_{n_{e}},\mathbf{\hat{V}}^{d}_{pred}\}$.

    \item \textbf{InstructICT-$\text{E}_{\text{T}}$}: To explore the performance of adopting direct interactions between the visual features and the textual features, we replace the learned queries with the text inputs. the formulation of InstructICT-$\text{E}_{\text{T}}$ can be symbolized as $\{\mathbf{\hat{T}}^{e}_{1},...,\mathbf{\hat{T}}^{e}_{n_{e}},\mathbf{\hat{V}}^{d}_{pred}\}$.

\end{itemize}

To compare and analyze these in-context demonstration variants more efficiently, we randomly sample one third data of the datasets for using in the ablation studies. 
We show the ablation study results of these variants over four different datasets in Tab.~\ref{tb:ab_ICT}.
From the results, we have the following findings:
\begin{enumerate} 

\item When we replace the visual-guided textual features $\mathbf{\hat{T}}^{c}$ from demonstration examples with the uni-modal textual features $\mathbf{T}^{b}$, i.e., transitioning from \ours to VanillaICT-$\text{E}_{\text{T}}$, we observe that the performance remains almost unchanged on image captioning and video captioning tasks, while it declines on VQA and VideoQA tasks. The observation suggests that information redundancy exists across different modalities in captioning tasks, whereas multi-modal information is crucial for visual/video question answering tasks.

\item InstructICT-$\text{E}_{\text{T}}$ exhibits comparable performance on image captioning and VQA tasks when compared to \ours. However, its performance is inferior to \ours on video captioning and VideoQA tasks. The difference between InstructICT-$\text{E}_{\text{T}}$ and \ours lies in the fact that textual features directly interact with visual features through cross-attention in the former, while textual features and learned queries interact with each other through self-attention in the latter. A possible explanation for the performance difference could be that the information contained in images is mostly useful, whereas video clips contain redundant information. On the other hand, learned queries demonstrate a strong ability to extract useful information, which can alleviate this issue.

\item To illustrate the existence of a modality gap between vision and text, we compare the performance of using visual-guided textual features and textual-guided visual features as demonstrations, i.e., \ours and InstructICT-$\text{E}_{\text{V}}$. And significant performance gaps are observed between them. Furthermore, InstructICT-$\text{E}_{\text{V}}$ is almost inferior to all other variants. This observation indicates that MM-LLMs may struggle to learn from in-context visual features and could even be misled by them.

\item VanillaICT-$\text{B}_{\text{VT}}$ and VanillaICT-$\text{B}_{\text{T}}$ directly feed raw text into the LLM. Different from them, InstructICT-$\text{E}_{\text{VT}}$ and VanillaICT-$\text{E}_{\text{T}}$ firstly employ M-Hub to encode raw text, and then input the enhanced representations into the LLM. However, no performance improvements are observed between them. We suspect that the powerful understanding ability of LLMs causes them to overlook the potential benefits of M-Hub.

\end{enumerate}

\begin{table}[t]
\centering
\caption{Performance using different numbers of demonstration examples.}
\resizebox{0.27\textwidth}{!}{
\begin{tabular}{cccc}
\toprule
\multirow{3}{*}{$n_{e}$} & \multicolumn{3}{c}{MSVD}                       \\ \cmidrule(r){2-4} 
                         & \multicolumn{2}{c}{Caption}    & VideoQA       \\ \cmidrule(r){2-3} \cmidrule(r){4-4}
                         & B@4           & C              & Acc          \\ \midrule
0                        & 75.1          & 176.9          & 64.9          \\
1                        & 75.2          & 177.1          & 65.9          \\
2                        & \textbf{76.5} & 179.1          & \textbf{66.7} \\
3                        & 76.1          & \textbf{179.5} & 64.7          \\ 
4                        & 75.4          & 178.7          & 64.7  \\ \bottomrule
\end{tabular}
}
\label{tb:ab_ne}
\end{table}

\begin{table}[t]
\centering
\caption{Performance using different sample strategies for demonstration examples.}
\resizebox{0.6\columnwidth}{!}{
\begin{tabular}{ccccccc}
\toprule
\multirow{3}{*}{Sample strategy} & \multicolumn{4}{c}{Caption}                         & VQA   & VideoQA \\ \cmidrule(r){2-7} 
                                 & \multicolumn{2}{c}{COCO} & \multicolumn{2}{c}{MSVD} & VQAv2 & MSVD    \\ \cmidrule(r){2-3} \cmidrule(r){4-5} \cmidrule(r){6-6} \cmidrule(r){7-7}
                                 & B@4        & C           & B@4        & C           & Acc   & Acc    \\ \midrule
Random                           & \textbf{43.6}       & \textbf{145.7}       & \textbf{76.5}       & \textbf{179.1}       & 72.3  & \textbf{66.7}    \\
One-to-many                      & \textbf{43.6}       & 145.2       & 76.2       & 178.3       & \textbf{72.6}  & 66.1    \\ \bottomrule
\end{tabular}
}
\label{tb:ab_sample}
\end{table}

\subsection{The Impacts of Sampling for Demonstrations}

We also investigate the impacts of different settings of sampling for demonstrations on the performance.

\subsubsection{Number of Samples} 

Tab.~\ref{tb:ab_ne} provides the experimental results with the sample number $n_{e}$ varying from 0 to 4 on MSVD. Note that when $n_{e}=0$, the model generates outputs only according to the textual-guided visual features from query inputs. From Tab.~\ref{tb:ab_ne}, it is observable that \ours can effectively learn from in-context information when $n_{e}$ is set to 1 or 2. However, as $n_{e}$ continues to increase, the performance remains almost unchanged and may even decline. The observation suggests that the MM-LLMs could be negatively influenced by the in-context information when $n_{e}$ becomes overly large.

\subsubsection{Sampling Strategy} 

The performance using different sampling strategies for demonstration examples is reported in Tab.~\ref{tb:ab_sample}. 
One-to-many indicates that for each video-text pair in the dataset, we randomly sample its demonstration examples from the same video that have different text paired with them. 
The results in Tab.~\ref{tb:ab_sample} show that \ours is robust to changes in the different sample strategies.

\subsection{The Impacts of Demonstrations on Inference}

\begin{table}[t]
\centering
\caption{Results of using demonstrations during inference.}
\label{tb:ict}
\scalebox{0.8}{
\begin{tabular}{@{}cccccccc@{}}
\toprule
\multirow{3}{*}{Method}           & \multirow{3}{*}{Demonstrations}          & \multicolumn{4}{c}{Caption}                         & VQA   & VideoQA \\ \cmidrule(r){3-6} \cmidrule(r){7-7} \cmidrule(r){8-8}
                                  &                                     & \multicolumn{2}{c}{COCO} & \multicolumn{2}{c}{MSVD} & VQAv2 & MSVD    \\ \cmidrule(r){3-4} \cmidrule(r){5-6} \cmidrule(r){7-7} \cmidrule(r){8-8} 
                                  &                                     & B@4        & C           & B@4        & C           & Acc   & Acc     \\ \midrule
VanillaFT                         & \multirow{3}{*}{\textcolor{red}{\XSolidBrush}} & 42.4       & 144.5       & 73.0       & 174.3       & 69.6  & 63.9    \\
VanillaICT-$\text{B}_{\text{VT}}$ &                                     & 42.4       & 142.5       & 75.4       & 178.5       & 69.8  & 64.3    \\
\ours                             &                                     & \textbf{43.6}       & 145.7       & \textbf{76.5}       & 179.1       & 72.3  & \textbf{66.7}    \\ \midrule
VanillaFT                         & \multirow{3}{*}{\textcolor{green}{\Checkmark}}   & 33.8       & 111.2       & 51.1       & 119.0       & 67.0  & 47.8    \\
VanillaICT-$\text{B}_{\text{VT}}$ &                                     & 43.2       & 143.8       & 75.3       & 178.2       & 70.0  & 56.6    \\
\ours                             &                                     & \textbf{43.6}       & \textbf{145.9}       & \textbf{76.5}       & \textbf{180.0}       & \textbf{72.8}  & 56.7    \\ \bottomrule
\end{tabular}
}
\end{table}

\begin{table}[t]
\centering
\caption{Results using different levels of features in VanillaFT.}
\scalebox{0.9}{
\begin{tabular}{@{}ccccccc@{}}
\toprule
\multirow{3}{*}{Level} & \multicolumn{4}{c}{Caption}                                    & \multicolumn{2}{c}{VideoQA}   \\ \cmidrule(r){2-5} \cmidrule(r){6-7}
                       & \multicolumn{2}{c}{MSR-VTT}   & \multicolumn{2}{c}{MSVD}       & MSR-VTT       & MSVD          \\ \cmidrule(r){2-3} \cmidrule(r){4-5} \cmidrule(r){6-6} \cmidrule(r){7-7} 
                       & B@4           & C             & B@4           & C              & Acc           & Acc           \\ \midrule
Frame                  & \textbf{51.3} & 74.7          & 73.0          & 174.3          & \textbf{43.4} & \textbf{63.9} \\
Video                  & 51.0          & \textbf{75.1} & \textbf{74.6} & \textbf{177.9} & 42.9          & 63.3          \\ \bottomrule
\end{tabular}
}
\label{tb:feat}
\end{table}

Tab.~\ref{tb:ict} presents the impact of using demonstrations on the model's performance during the inference phase. 
The used LLM is FlanT5.
Note that the results for the default setting reported in Tab.~1 of our submission use demonstrations during fine-tuning.
For each data sample in the test set, we randomly sample its demonstrations from the training set.
The results lead us to the following observations:
\begin{enumerate}
    
    \item The performance of \ours gets slightly improved when demonstrations are incorporated during inference across most tasks. However, for the VideoQA task on the MSVD dataset, demonstrations appear to affect the performance of all methods negatively. Despite this, \ours consistently surpasses baselines.

    \item VanillaICT-$\text{B}_{\text{VT}}$ demonstrates similar performance on most tasks, regardless of whether demonstrations are used during the inference phase or not. In contrast, a significant decline in performance can be observed for VanillaFT when demonstrations are incorporated. 
    
\end{enumerate}

Above findings indicate that the use of demonstrations during inference can have varying effects on different methods and tasks.
\ours and VanillaICT-$\text{B}_{\text{VT}}$ have fully learned from demonstrations during in-context tuning and they show comparable performance regardless of whether demonstrations are leveraged during inference.
Besides, the observations underscore the importance of in-context tuning in enabling MM-LLMs to learn from demonstrations effectively.

\subsection{The Impacts of Feature Levels}

Tab.~\ref{tb:feat} shows the performance of using different levels of features in VanillaFT. 
For frame-level features, we pass each video clip through the frozen image encoder and M-Hub individually, and then concatenate them together to obtain frame-level features. 
The results in Tab.~\ref{tb:feat} illustrate that, despite using 16 times fewer tokens (i.e., frame number $n_{f}$), the performance using video-level features is comparable to, and in some cases even surpasses, the performance achieved using frame-level features. 
The observation further demonstrates that video clips contain redundant information which does not significantly contribute to the performance. 
Note that, due to the limitations of the input length of LLMs, we do not conduct experiments for VanillaICT-$\text{B}_{\text{VT}}$ and \ours, which additionally take demonstrations as input.

\begin{figure*}[t]
    \centering
    \includegraphics[width=1\textwidth]{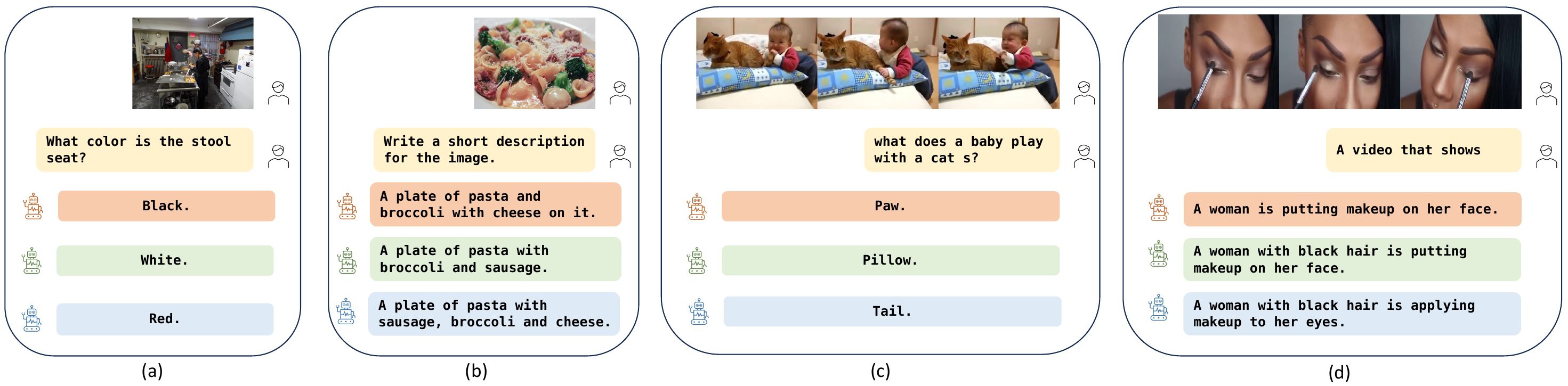}
    \caption{Case study on (a) visual question answering, (b) image captioning, (c) video question answering, and (d) video captioning. We show the answers generated by VanillaFT, VanillaICT-$\text{B}_{\text{VT}}$ and \ours in orange, green and blue, respectively.}
    \label{fig:cs}
\end{figure*}

\begin{table}[t]
\centering
\caption{Results on different sets of VQAv2.}
\label{tb:vqav2}
\resizebox{0.37\textwidth}{!}{
\begin{tabular}{@{}cccc@{}}
\toprule
\multirow{2}{*}{Method}           & \multirow{2}{*}{LLM}    & \multicolumn{2}{c}{VQAv2}     \\ \cmidrule(l){3-4} 
                                  &                         & val           & test-dev      \\ \midrule
VanillaFT                         & \multirow{3}{*}{FlanT5} & 69.6          & 71.6          \\
VanillaICT-$\text{B}_{\text{VT}}$ &                         & 69.8          & 71.8          \\
\ours     &                         & \textbf{72.3} & \textbf{74.6} \\ \midrule
VanillaFT                         & \multirow{3}{*}{OPT}    & 56.9          & 60.0          \\
VanillaICT-$\text{B}_{\text{VT}}$ &                         & 62.7          & 64.0          \\
\ours     &                         & \textbf{73.0} & \textbf{75.2} \\ \bottomrule
\end{tabular}
}
\end{table}

\subsection{Case Study}

We randomly sample some cases covering four different multi-modal downstream tasks.
The results of the sampled cases generated by \ours, VanillaFT and VanillaICT-$\text{B}_{\text{VT}}$ are shown in Fig.~\ref{fig:cs}. 
From Fig.~\ref{fig:cs}, we can observe that \ours is capable of better understanding the detailed information contained in multi-modal data, while it is difficult for baselines to capture the details.
For instance, in Fig.~\ref{fig:cs}(d), baselines only recognize the face while \ours can identify that the target is eyes.

\section{Experiments on VQAv2}

Note that the test labels of VQAv2 dataset are not publicly available.
Tab.~\ref{tb:vqav2} reports the results of \ours compared with two baselines, i.e., VanillaFT and VanillaICT-$\text{B}_{\text{VT}}$, on the validation set and the test set. 
The results demonstrate that \ours consistently outperforms the baselines on the validation set and the test set of VQAv2.

\section{The differences between \ours and Otter}
\label{cmp_otter}

Otter is the most similar wrok compared with \ours, we have briefly discussed it in Sec.~\ref{rw_icl}. In this section, we provide detailed comparisons between them, including the experimental results.

As shown in Tab.~\ref{tab:diff_otter}, the four main differences between \ours and Otter are as follows:

\begin{enumerate}
    \item Otter uses in-context examples during pre-training, while MMICT is a novel fine-tuning paradigm applied to downstream tasks.

    \item The interaction of visual and textual features is implemented in M-hub. Conversely, Otter incorporates additional cross-attention modules within the layers of LLM, leading to an increase in computational requirements compared to M-hub.

    \item The data format of Otter necessitates meticulous design, as depicted in Fig.~2 of their paper. In contrast, our data format is more flexible and does not mandate a specific design. Remarkably, our method can function effectively even with random sampling for demonstrations, as illustrated in Section 4.7.

    \item At inference, Otter generates predictions based on the provided in-context examples. Differently, MMICT exhibits robust performance without demonstrations.
\end{enumerate}

\begin{table}[]
\centering
\caption{The differences to Otter.}
\label{tab:diff_otter}
\resizebox{0.6\columnwidth}{!}{
\begin{tabular}{ccccc}
\hline
\multirow{2}{*}{Method} & \multirow{2}{*}{\begin{tabular}[c]{@{}c@{}}Target\\  Stage\end{tabular}} & \multirow{2}{*}{\begin{tabular}[c]{@{}c@{}}Multi-modal\\  Fusion\end{tabular}} & \multirow{2}{*}{\begin{tabular}[c]{@{}c@{}}In-context \\ Unrestriction\end{tabular}} & \multirow{2}{*}{\begin{tabular}[c]{@{}c@{}}Prediction \\ w/o Examples\end{tabular}}             \\ 
                        &                                                                          &                                                                               &                                                                                               \\ \hline
Otter                   & Pre-training   & LLM                                                          & \textcolor{red}{\XSolidBrush}                                                              & \textcolor{red}{\XSolidBrush}                                                                       \\
MMICT                   & Fine-tuning    & M-Hub                                                          & \textcolor{green}{\Checkmark}                                                         & \textcolor{green}{\Checkmark}                                                                     \\ \hline
\end{tabular}
}
\end{table}

\begin{table}[]
\centering
\caption{Results compared with Otter on MSVD tasks.}
\label{tab:cmp_otter}
\resizebox{0.42\columnwidth}{!}{
\begin{tabular}{@{}clccc@{}}
\toprule
\multirow{2}{*}{Method} & \multirow{2}{*}{LLM} & \multicolumn{2}{c}{Caption}    & VideoQA            \\ \cmidrule(l){3-5} 
                        &                      & B@4            & C              & Acc           \\ \midrule
Otter                   & LLaMa7B              & 78.9          & \textbf{184.2} & 65.2          \\
\ours                   & OPT2.7B              & \textbf{80.4} & 180.4          & \textbf{66.3} \\ \bottomrule
\end{tabular}
}
\end{table}

Furthermore, as displayed in Tab.~\ref{tab:cmp_otter}, we also provide a direct comparison with Otter. From the results, we can find that although the size of LLM in \ours is much smaller than that in Otter, \ours exists comparable and even superior performance on video captioning and VideoQA tasks when compared to Otter. 
\section{Conclusion}
\label{sec:con}

In this paper, we propose \ours for boosting multi-modal fine-tuning with in-context examples. 
\ours enables MM-LLMs to learn from visual-guided textual features of demonstrations, and subsequently generate outputs with the textual-guided visual features of input queries.
We propose the M-Hub used in \ours to capture the multi-modal fused features within a unified architecture.
Furthermore, we design various demonstration variants by fully considering the flexibility of M-Hub.
From our extensive experiments conducted across six different multi-modal datasets, we can find that \ours exceeds traditional fine-tuning strategy and VanillaICT.
Additional experiments on different demonstration factors (i.e., feature extraction strategy, sampling number for demonstrations and sampling strategies for demonstrations) further ascertain the effectiveness and robustness of \ours. 
In the future, we plan to experiment \ours with more modalities (e.g., audio) and verify its effectiveness on more multi-modal tasks.

\section*{Acknowledgments}
This work is supported by National Science and Technology Major Project (No. 2022ZD0118201) and National Natural Science Foundation of China (No. 62002303, 42171456).





\end{document}